\documentclass[9.5pt,journal,twoside,twocolumn]{IEEEtran}

\pdfoutput=1
\usepackage{graphicx}
\usepackage{balance}
\usepackage{math}

\begin{document}
\bibliographystyle{IEEEtrans}
\title{Visually Guided Object Grasping\thanks{The work described herein
has been supported by the European ESPRIT-III programme through the SECOND
project (Esprit-BRA No. 6769).}}
\author{Radu Horaud, {\em Member, IEEE}\thanks{R. Horaud is with
GRAVIR-CNRS and INRIA Rh\^one-Alpes, 655 avenue de
l'Europe, 38330 Montbonnot France}, 
Fadi Dornaika\thanks{F. Dornaika is with Department of Mechanical and Automation Engineering The Chinese University of Hong Kong Shatin, NT, Hong Kong.},
and Bernard Espiau\thanks{B. Espiau is with INRIA Rh\^one-Alpes}\\
}
\maketitle
\begin{abstract}
In this paper we present a visual servoing approach to the problem of object
grasping and more generally, to the problem of aligning an end-effector with
an object. 
First we extend the method proposed in
\cite{EspiauChaumetteRives92} to the case of a camera which is not mounted
onto the robot being controlled 
and we stress the importance of the
real-time estimation of the image Jacobian. 
Second, we show how to represent
a grasp or more generally, an alignment between two solids in 3-D projective
space using an uncalibrated stereo rig. Such a 3-D projective representation
is view-invariant in the sense that it can be easily mapped into an 
{\em image set-point} without any knowledge about the camera parameters. 
Third, we perform an analysis of the performances of
the visual servoing algorithm and of the grasping precision that can be
expected from this type of approach.

\end{abstract}
\begin{keywords} Object grasping, 
projective camera model, 3-D projective reconstruction, hand-eye coordination,
visual servoing.
\end{keywords}

\section{Introduction}
\label{section:Introduction}

One of the most common tasks in robotics is grasping. Although the
importance of grasping has been recognized for many years, there are only
a few grasping systems that can operate in complex environments. This is
mainly due to the difficulty to execute precise robot hand motions
in the presence of
various perturbations: the robot's kinematic is known only partially,
unpredictable obstacles may be located in the neighborhood of the object to be
grasped, and
the location of the object to be grasped with respect to the robot
may be either poorly known or not known at all.

Our approach to perform automatic grasping follows the classical approach of
splitting the task in off-line and on-line stages. 
The goal of the off-line stage is to select a grasp -- specify a relationship
between the gripper and the object -- and represent this relationship in some
space. The task to be achieved on-line is to control the robot's motion
such that the gripper moves from its initial position to a final position that
is consistent  with the planned grasp. With our approach both off- and on-line stages use
cameras, therefore intrinsic and extrinsic camera calibration will affect the
behavior of the grasping process and the accuracy with which the grasping
location will eventually be reached. Hence, one of the most important merits of
a visually guided grasping technique is to be robust with respect to 
internal and external camera parameters. Alternatively, one may devise a
method which uses uncalibrated cameras.

Consider, for example, the following scenario. The off-line stage -- which may
well be viewed as a preparation or planning stage -- takes place in a
laboratory. The on-line stage -- task execution -- takes place in a hazardous
or remote site (nuclear, space, offshore, etc.). The cameras 
used in the laboratory are not the same as the remote cameras.
Moreover, the locations (position and orientation) of the cameras with respect
to the object to be grasped and with respect to the robot are not the same in
the laboratory and remote site.

In this paper we develop a visual servoing based method that is able of
achieving grasping or, more generally, alignment tasks. The main feature of the
method described herein is that the accuracy associated with the task to be
performed is not affected by discrepancies between
the Euclidean setups at task preparation and at task execution stages. By
Euclidean setup we mean internal camera calibration and camera-to-world and
robot-to-world relationships.

More precisely, the desired object to gripper alignment will be represented in
3-D projective space rather than in 3-D metric space. Such a non-metric
representation can be obtained with an uncalibrated pair of cameras, or a
stereo rig. During the off-line stage one stereo rig observes both the object
and the gripper in their aligned setup and performs a projective
reconstruction of both of them. During the on-line stage another stereo rig
observes the object and performs its projective reconstruction. Hence, two
projective reconstructions of the object are available in two different
projective bases, each one of these bases being attached to each one of the
two stereo rigs. Therefore it is possible to compute a 3-D projective
transformation between the off-line and on-line setups, transfer the
gripper from one setup to another, and predict the location of the gripper in
the images associated with the second stereo rig. Once this off-line to
on-line transfer of gripper points from one image pair to another image pair
has been performed, the problem of moving the gripper from an initial position
to the desired grasp position becomes a classical image-based robot servoing
problem: 
\begin{enumerate} 
\item estimate the velocity screw associated with the gripper frame;
\item move the robot until the image points associated with the observed gripper
are properly aligned with their predicted locations.
\end{enumerate}

The visual grasping scheme that we just described suggests that:
\begin{enumerate}
 \item two
cameras are involved in the visually guided control loop;
\item these
cameras must be calibrated \cite{HutchinsonHagerCorke96}. 
\end{enumerate}

In fact, once the
gripper points have been properly transferred, the visual servoing process can
proceed with only one of the two cameras and hence, only one among these two
cameras must be internally calibrated. Recently it has been shown by one of us
that internal camera calibration weakly affects the convergence of image-based
robot control when only one camera is being used \cite{Espiau93}.

\section{Background, contribution, and paper organization}
The theory of image-based servoing has been developed, in parallel, by a
number of researchers \cite{EspiauChaumetteRives92}, \cite{Hashimoto91},
\cite{Maru93}, \cite{Hager94a},
\cite{Feddema93}, \cite{Corke93a}, \cite{Corke93b}, \cite{Sharma94}.
Central to the image-based approach is the necessity to compute the
image Jacobian. This is equivalent to computing the differential
relationship between a
scene frame and the camera frame (either the scene or the camera frame is
attached to the robot). Jacobian estimation requires knowledge about the
camera intrinsic and extrinsic parameters. The latter parameters amount to the
rigid mapping between the scene frame and the camera frame. Many
implementations get around this problem by simply
allocating constant values to the image Jacobian.

The debate whether the sensor should be mounted onto the robot (eye-in-hand)
or should be mounted onto a fixture (independent-eye) is important because
each one of these two setups has limitations and advantages. With a hand-eye
approach, the setup (camera parameters and hand-eye relationship) at planning
must be identical with the setup at runtime. The independent eye approach
offers more flexibility at the price of the use of several cameras rather than
a single camera.

This paper has the following contributions. In
section~\ref{section:Vision-based-robot-motion}
we extend the hand-eye servoing method proposed in
\cite{EspiauChaumetteRives92} to the independent-eye setup. 
Within the context of the new mathematical expression
that we derive for the image Jacobian, we make clear which parameters vary
with time and which parameters remain constant. Indeed, in a recent review
paper \cite{HutchinsonHagerCorke96} this analysis was not available.
Moreover we stress the importance of on-line
pose computation. 

In
section~\ref{section:View-invariant} we show how to represent an alignment
between two objects in 3-D projective space. The alignment condition thus
derived is projective invariant in the sense that it can be used in
conjunction
with
two uncalibrated camera pairs (one at planning and one at runtime) to compute
a goal position for visual servoing.

In sections~\ref{section:performance-analysis} and
\ref{section:grasping-experiments}
we describe an in depth comparison of image-based servoing with a fixed
(approximated) Jacobian and with a variable (exact) Jacobian. Next
we describe the
implementation of a visually-guided grasping
system which
integrates the results of
sections~\ref{section:Vision-based-robot-motion},
\ref{section:View-invariant},
together with a pose computation
method. Finally, section~\ref{section:Discussion} gives some
directions for future work.

\section{Image-based servoing}
\label{section:Vision-based-robot-motion}

In this section we consider a camera that observes a moving robot gripper. First
we determine the image Jacobian associated with such a configuration. Second
we define a visual servoing process that allows the camera to control the robot
motion such that the gripper reaches a previously determined image set
position --
one way to compute such an image set position using an uncalibrated stereo rig
will be described in section~\ref{section:View-invariant}.

\subsection{Image Jacobian}

\begin{figure}[t!]
\centering
\includegraphics[width=0.49\textwidth]{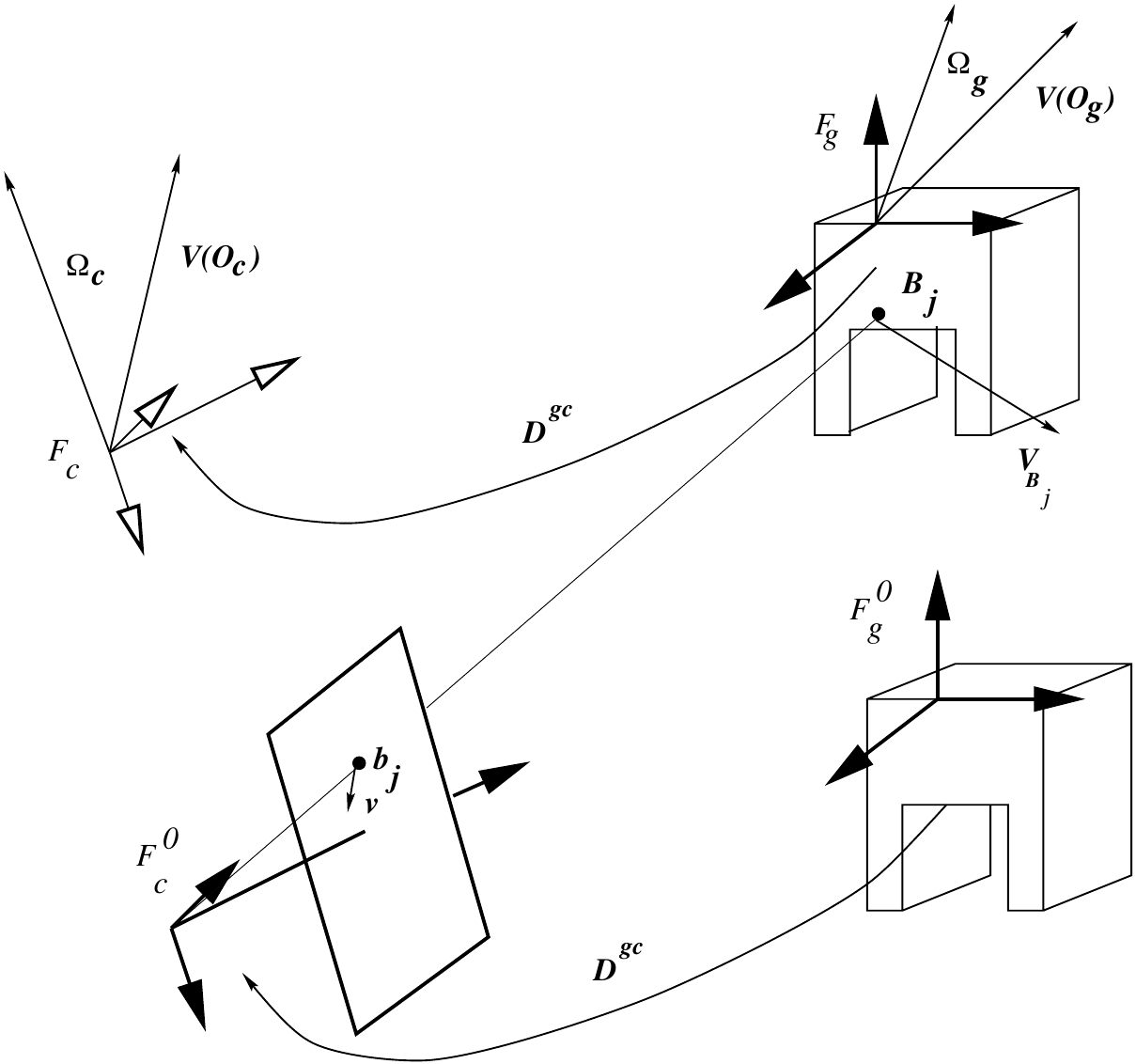}
\caption{This figure shows the relationships between the various frames 
associated with the gripper and with the camera.}
\label{fig:torseur}
\end{figure}

Let us define two useful Euclidean frames as follows, Figure~\ref{fig:torseur}:
\begin{enumerate}
\item $F_g^0$ is the gripper reference frame associated with
the gripper in its
initial position prior to visual servoing and
\item $F_c^0$ is the camera reference frame; since the camera will remain fixed
while the gripper will move, the frame attached to the camera is a fixed 
reference frame.
\end{enumerate}
Let $\Dmat^{gc}$ be the 4$\times$4 homogeneous matrix mapping $F_g^0$ onto
$F_c^0$. Next we consider the gripper while it moves and we define two moving
frames rigidly attached to the gripper:
\begin{enumerate}
\item $F_g$ which is a moving gripper frame related to $F_g^0$ by the 
continuous displacement $\Dvect^g(t):F_g^0 \rightarrow F_g$ and
\item $F_c$ which as a moving frame as well rigidly attached to the gripper
related to $F_c^0$ by the continous displacement
$\Dvect^c(t):F_c^0 \rightarrow F_c$.
\end{enumerate}
Clearly the homogeneous matrix mapping $F_g$ onto $F_c$ is the same as the
matrix mapping $F_g^0$ onto $F_c^0$ and is equal to:
\begin{equation}
        \Dmat^{gc} = \left( \begin{array}{cc}
        \Rmat^{gc} & \tvect^{gc} \\
        \zerovect\tp & 1
        \end{array} \right)
\label{eq:displacement}
\end{equation}

At each time $t$ the two displacements $\Dvect^g(t)$ and $\Dvect^c(t)$ are
conjugated:
\[	\Dvect^c(t) = (\Dmat^{gc})\inverse \; \Dvect^g(t) \; \Dmat^{gc}	\]

Consequently the motion of the gripper can be expressed either by the moving
frame $F_g$ with respect to $F_g^0$
or by the moving frame $F_c$ with respect to $F_c^0$: 
\[ \Tvect_g=\{\Vvect (O_g), \Omega _g\} \]
is the velocity screw
of $F_g$ with respect to $F_g^0$ and 
\[ \Tvect_c=\{\Vvect (O_c), \Omega _c\} \]
is
the velocity screw 
of $F_c$ with respect to $F_c^0$. These two screws are related by the formula:
\begin{equation}
\Tvect_c = \Theta^{gc} \Tvect_g
\label{eq:velocity-mapping}
\end{equation}
with:
\begin{equation}
\Theta^{gc} = \left( \begin{array}{cc}
\Rmat^{gc} & \Rmat^{gc} S(\tvect^{gc}) \\
\zerovect      & \Rmat^{gc}
\end{array} \right)
\label{eq:Theta-definition}
\end{equation}
where
$S(\avect)$ is the skew-symmetric matrix associated with a 3-vector $\avect$.
It is important to notice that the rotation $\Rmat^{gc}$ and translation
$\tvect^{gc}$ describe the {\em initial} pose of the gripper with respect to
the camera and hence they remain constant during visual servoing.

Now, let $B_j$ be a 3-D point onto the gripper and let $\Bvect_j^c=(x_j,
y_j,z_j)\tp$ be its Euclidean coordinates in the camera-centered  frame 
$F_c^0$, e.g., figure~\ref{fig:torseur}. The
projection of this point onto the image has as coordinates:
\begin{eqnarray}
\label{eq:u-hand-projection}
u_j & = & \alpha _u \frac{x_j}{z_j} + u_0 \\
\label{eq:v-hand-projection}
v_j & = & \alpha _v \frac{y_j}{z_j} + v_0
\end{eqnarray}
where $\alpha _u$, $\alpha _v$, $u_0$, and $v_0$ are the well known intrinsic
camera parameters associated with a pin-hole model and $(u,v)$ are the image
coordinates of a pixel. By computing the time
derivatives of $u_j$ and $v_j$ in equations (\ref{eq:u-hand-projection}) and
(\ref{eq:v-hand-projection}),
knowing that
$\stackrel{\bullet}{\Bvect_j^c} = \Vvect (O_c) + \Omega _c \times \Bvect_j^c$,
and by combining with eq.~(\ref{eq:Theta-definition}), it
is
straightforward to
obtain:
\begin{equation}
\left( \begin{array}{c}
\stackrel{\bullet}{u}_j \\ \stackrel{\bullet}{v}_j
 \end{array}\right)
=
\Jmat_j
\Tvect_{g}
\label{eq:jacobian-mapping}
\end{equation}
with $\Jmat_j= \Lmat_j \Theta ^{gc}$ and $\Lmat_j$ equal
to:
\[
\left( \begin{array}{cc} \alpha_u & 0 \\ 0 & \alpha_v \end{array} \right)
\left( \begin{array}{cccccc}
\frac{1}{z_j} & 0 &  \frac{-x_j}{z_j^2} & \frac{-x_jy_j}{z_j^2} &
1+\frac{x_j^2}
{z_j^2} &
\frac{-y_j}{z_j}\\
0 & \frac{1}{z_j} & \frac{-y_j}{z_j^2} & -1-\frac{y_j^2}{z_j^2} &
\frac{x_jy_j}{
z_j^2} &
\frac{x_j}{z_j}
 \end{array}\right)
\]

\subsection{Control law}

As already mentioned, we consider $n$ 3-D points ($B_j$)
onto the robot gripper together with their projections onto the image 
($\bvect _j=(s\,u_j, \; s\,v_j, \; s)$).
Let $\svect$ be the image vector formed with the Euclidean
coordinates of all the
points $\bvect _j$. For $n$ points, the vector $\svect$ has 2$\times n$ components:
\[
\svect = \left( u_1 \; v_1 \ldots u_j \; v_j \ldots u_n \; v_n \right)\tp
\]
We denote by $\svect ^{\star}$ the image set-point -- 
the final (goal) position. This goal position may correspond, for example, to an
alignment condition for grasping (see section~\ref{section:View-invariant}) or to
any other goal position that one wants to reach.

Therefore, the task consists in moving the robot such that the Euclidean
norm of the error vector $\svect - \svect^{\star}$ decreases. 
Hence, one may constrain the image velocity of
each point being considered to exponentially reach its goal position with time. 
This desired behavior writes as
$\stackrel{\bullet}{\svect} = g \left( \svect ^{\star} - \svect \right)$
where $g$ is a positive scalar that controls the convergence rate
of the visual servoing.

It is now possible to combine the above formula with eq.~(\ref{eq:jacobian-mapping}) 
and we obtain:
\begin{equation}
\Jmat \Tvect_g
= g \left( \svect ^{\star} - \svect \right)
\label{eq:control-law}
\end{equation}
With
$\Jmat\tp = \left( \Jmat_1\tp \ldots \Jmat_n\tp \right)$.
Let us now assume that the rank of the $n\times$6 matrix $\Jmat$ is 6 (i.e $n \geq 3$, and the 
gripper points $B_j$ are not collinear).
The control velocity screw may then be 
computed as:
\begin{equation}
\Tvect_{g}
= g \left( \hat{\Jmat}\tp \Wmat \hat{\Jmat} \right)^{-1} \hat{\Jmat}\tp W\left( \svect ^{\star} - \svect 
\right) = g \hat{\Jmat}^\dag \left( \svect ^{\star} - \svect
 \right)
\label{eq:control-velocity}
\end{equation}
where $\Wmat$ is a symmetric positive matrix of rank 6 allowing, for example, to select some 
preferred points in the image among the $n$ points that are available, 
and $\hat{\Jmat}$ is the model
of $\Jmat$ which is used in the control expression.

To compute this model $\hat{\Jmat}$, it is therefore necessary to estimate 
the constant matrix
$\Theta ^{gc}$ and the time-varying values of $x_j$, $y_j$, and
$z_j$ in $F_c$. 

Let $\Bvect_j^g$ be the coordinates of a gripper point in the gripper frame
$F_g$. These coordinates can be easily estimated off-line using a hand-tool 
calibration technique and which is described in \cite{DornaikaHoraud98}. 
In order to estimate
the initial pose of the gripper with respect to the camera, i.e.,
$\Dmat^{gc}$, one has to apply a pose computation method to a set of 2-D to
3-D point matches $\bvect_j\leftrightarrow \Bvect_j^g$ when the gripper is in
its initial position. 
Moreover, $x_j$, $y_j$, $z_j$ -- the camera coordinates of $B_j$ can also be evaluated through a pose computation method, the pose method being applied at each
time to the matches $\bvect_j\leftrightarrow \Bvect_j^g$. 

Pose computation is a classical
problem in computer vision 
and photogrammetry
and many closed-form and/or numerical
solutions have been proposed in the past.
Nevertheless, these solutions to the
object pose computation problem were not entirely satisfactory. This is the
main reason for which the current solution used in visual servoing consists in
considering that the pose parameters do not vary too much over time and hence
$\hat{\Jmat}$ is often obtained by giving to the entries of $\Jmat$ constant values, for example  
those corresponding to the goal position
\cite{EspiauChaumetteRives92}. Even if the stability of the closed loop system can be preserved
 as long as $\Jmat \hat{\Jmat}^\dag$ is a positive matrix, the
convergence can nevertheless be strongly affected.
In \cite{HoraudDornaikaLamiroyChristy97} we present a new
object pose computation method that is fast and reliable enough to be incorporated
in the real-time loop of the visual servoing algorithm and it will be shown in the following
that its performances will be significantly improved compared to the classical approach.

\section{Projective invariant object/gripper alignment}
\label{section:View-invariant}
The visual servoing method described in the previous section requires knowledge
of the set-point $\svect^{\star}$ which is a set of image points. 
$\svect ^{\star}$ is a function of the camera/gripper relationship (extrinsic
parameters) and of the camera internal model (intrinsic parameters). 
Whenever the location of the
object to be grasped varies with respect to the camera, the set-point
$\svect^{\star}$ varies as well. In this section we show how to compute the
set-point $\svect^{\star}$ such that it is ``view-invariant", i.e., it is
independent of both intrinsic and extrinsic camera parameters. This will allow
more flexibility because the setups at learning and runtime stages can 
be different. 

In Euclidean space, 
the relationship between two objects is usually represented by
some rigid transformation. Alternatively, the object-gripper alignment, or any other
object-to-object relationship, can be represented in terms of relationships
between objects
points. The choice of these points depends upon the visual sensor being
used and hence upon the visual process allowing to extract image points, i.e.,
feature extraction. They are not
necessarily contact points
between the object and the gripper. Therefore they may not be present in the CAD
descriptions of both the object and the gripper.
The idea of our
approach is to represent such object-gripper relationships projectively: 3-D
object and gripper points are described into an object-centered projective
basis. 

\begin{figure}[t!]
\centering
\includegraphics[width=0.5\textwidth]{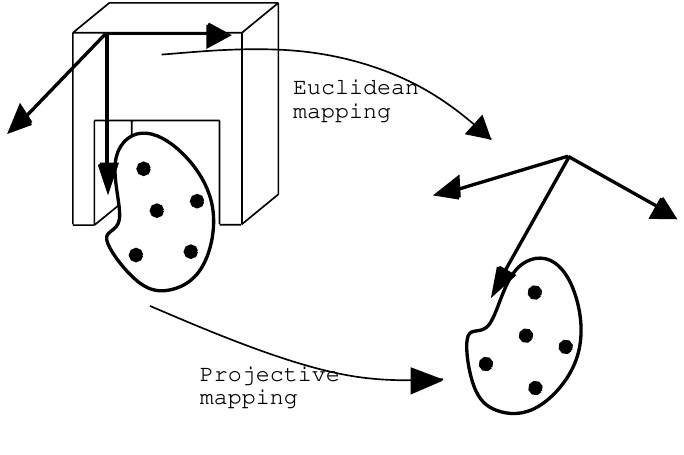}
\caption{The object projective basis is rigidly attached to the gripper
Euclidean frame. Whenever the object moves, these two frames remain virtually
attached to it. The projective mapping is conjugated to the Euclidean
mapping.}
\label{fig:gripper-motion}
\end{figure}

More precisely, consider an object to be grasped and a gripper aligned with
this object.
Let $A_i,\; i=1\ldots m$ be a set of 3-D
object points and $B_j,\;j=1\ldots n$ be a set of 3-D
gripper points. Among the object points consider five of them in general
position, say $A_1$ to $A_5$ (these five points form a basis
of the 3-D projective space) and let $\Avect^o_i$, $\Bvect^o_j$ be the
projective coordinates of the object and gripper points in this basis.
Moreover, consider an Euclidean frame attached to the gripper, $F_g$. 
Three points are
sufficient to uniquely define such an Euclidean frame. Notice that an
Euclidean frame is just a special case of a projective basis where one point
is the origin of the frame, three points on the plane at infinity correspond
to the directions of the three axes, and the fifth point defines the unit
vector \cite{SempleKneebone79}.

Therefore, the Euclidean space can be viewed as a subspace of the projective
space. There exists a projective transformation mapping Euclidean coordinates
onto projective coordinates. Such a transformation is conveniently described
by a 4$\times$4 invertible homogeneous
matrix and let $\Hmat^{go}$ be the matrix mapping
Euclidean coordinates onto projective coordinates from the gripper
frame onto the object basis described above. If we denote by $\Avect^e_i$,
$\Bvect^e_j$ the Euclidean coordinates of the points just mentioned we have:
\[	\Avect^o_i \simeq \Hmat^{go} \Avect^e_i \]
\[      \Bvect^o_j \simeq \Hmat^{go} \Bvect^e_j \]
where ``$\simeq$" denotes the projective equality.

Next we suppose that the object alone
lies in a different position and orientation.
Therefore, the object moved and since its motion is a rigid one it can be
described in the Euclidean frame mentioned above which remained virtually
linked to the object. Let $\Dmat$ be the rigid motion associated with the
object and with this particular frame. $\Dmat$ is a 4$\times$4 homogeneous
mapping of the form given by eq.~(\ref{eq:displacement}).
The equivalent {\em projective displacement} is (see
Figure~\ref{fig:gripper-motion}):
\[	\Hmat \simeq \Hmat^{go} \Dmat \left( \Hmat^{go} \right) \inverse \]
$\Hmat$ maps the ``old" projective coordinates into the ``new" ones and
$\Dmat$ maps the old Euclidean coordinates into the new ones but the
relationship between the Euclidean and projective representations of the
gripper-to-object alignment, $\Hmat^{go}$ 
remains invariant. In practice this representation
is encapsulated by the projective coordinates of gripper points in an object
centered projective basis: $\Bvect_1^o,\;\ldots \Bvect_n^o$ in the projective
basis $\Avect_1^o,\;\ldots \Avect_5^o$.

\subsection{Projective reconstruction with a camera pair}

\begin{figure}[t!]
\centering
\includegraphics[width=0.5\textwidth]{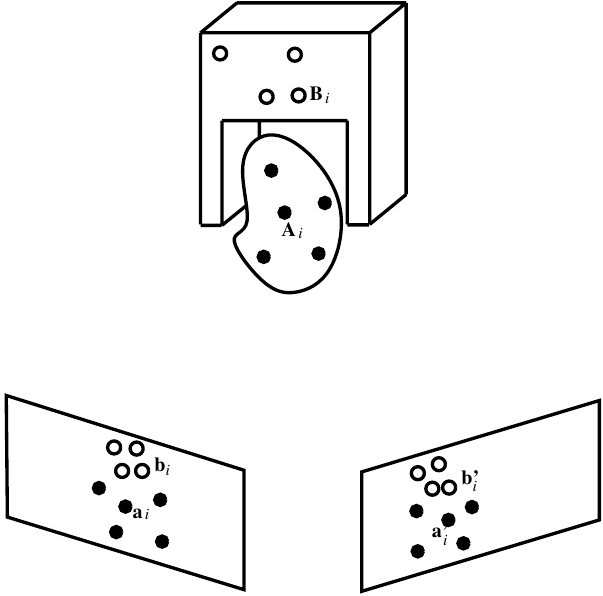}
\caption{The projective basis in which the camera pair reconstructs scene points is a sensor centered basis. Alternatively, one can select five points onto the
object to be grasped and built an object centered representation of the
gripper-to-object alignment.}
\label{fig:setpoint}
\end{figure}

We consider a pair of uncalibrated cameras which observe the gripper aligned with
the object, Figure~\ref{fig:setpoint}. 
It is known that from point-to-point matches between the two images it is
possible to compute the epipolar geometry associated with the two cameras
\cite{LuongFaugeras96}. Moreover, from the epipolar geometry 
two 3$\times$4 projection matrices mapping the 3-D projective
space onto the two images can be computed \cite{Hartley94c}. 
We denote by $\Pmat^x$ and $\Pmat'^x$ the two
projection matrices. Let $\mvect^x$ and $\mvect'^x$ be the projections of a
3-D point $M$ onto the left and right images associated with the two cameras.
The equations:
\begin{equation}
\mvect^x \simeq \Pmat^x \Mvect^x, \;\; \mvect'^x \simeq \Pmat'^x \Mvect^x
\label{eq:proj-rec-x}
\end{equation}
allow to compute the 3-D projective coordinates $\Mvect^x$ of the 3-D point
$M$ in a projective basis $x$ attached to the camera pair. Since the
geometry of the camera pair (intrinsic and extrinsic parameters) may change
over time, the camera pair is not a rigid object. However it is possible to
compute a projective transformation mapping the sensor centered projective
reconstruction $x$ into the object centered projective reconstruction $o$ just
described. For the sensor and object projective coordinates of a point $A_i$
we have:
\begin{equation}
\Avect^o_i \simeq \Hmat^{xo} \Avect^x_i
\label{eq:homography-xo}
\end{equation}
where $\Avect^x_i$ is obtained by applying eq.~(\ref{eq:proj-rec-x}) to an
object point being observed with the camera pair, and $\Hmat^{xo}$ is a
4$\times$4 projective transformation.

\subsection{Stereo point transfer}

At runtime, another stereo pair observes the object to be grasped. However,
the gripper is at some distance from the object and the task is to move the
gripper from its initial position to a {\em virtual position}. The latter
gripper position corresponds to the gripper-to-object alignment defined during
the off-line stage.

Let $\Pmat^y$ and $\Pmat'^y$ be the matrices associated with the runtime
camera pair $y$ and therefore we have:
\begin{equation}
\mvect^y \simeq \Pmat^y \Mvect^y, \;\; \mvect'^y \simeq \Pmat'^y \Mvect^y
\label{eq:proj-rec-y}
\end{equation}

Again, the sensor centered 3-D projective coordinates of an object point can
be mapped in a object centered description:
\begin{equation}
\Avect^o_i \simeq \Hmat^{yo} \Avect^y_i
\label{eq:homography-yo}
\end{equation}

By combining eqs.~(\ref{eq:homography-xo}) and (\ref{eq:homography-yo}) we
obtain a relationship between the projective coordinates of an object point
expressed in the two projective bases $x$ and $y$:
\begin{equation}
\Avect^y_i \simeq \left(\Hmat^{yo}\right)\inverse \Hmat^{xo} \Avect^x_i \simeq
\Hmat^{xy} \Avect^x_i
\label{eq:homography-xy}
\end{equation}

Eq. (\ref{eq:homography-xy}) allows to compute a 4$\times$4 homogeneous matrix
$\Hmat^{xy}$ from point matches between two setups, $x$ and $y$,
$(\avect^x,\avect'^x)\leftrightarrow (\avect^y,\avect'^y)$. With five
point matches 
one obtains an exact solution. However, if a larger number of point
matches are available, a least-square solution can be computed
\cite{HoraudCsurka98}. To summarize, the following procedure transfers gripper
points from the learning setup to the runtime setup:
\begin{enumerate}
\item For each gripper point $B_j,\;j=1\ldots n$:
\item Reconstruct the projective coordinates of a gripper point from its
images associated with the setup $x$:
\[
\bvect^x_j \simeq \Pmat^x \Bvect^x_j, \;\; \bvect'^x_j \simeq \Pmat'^x \Bvect^x_j
\]
\item Map these point coordinates from one projective basis to the other
projective basis:
\[ \Bvect^y_j \simeq \Hmat^{xy} \Bvect^x_j
\]
\item Project the gripper point onto the images associated with the runtime
setup:
\[
\bvect^y_j \simeq \Pmat^y \Bvect^y_j, \;\; \bvect'^y_j \simeq \Pmat'^y \Bvect^y_j
\]
\end{enumerate}

\begin{figure}
\centering
\includegraphics[width=0.5\textwidth]{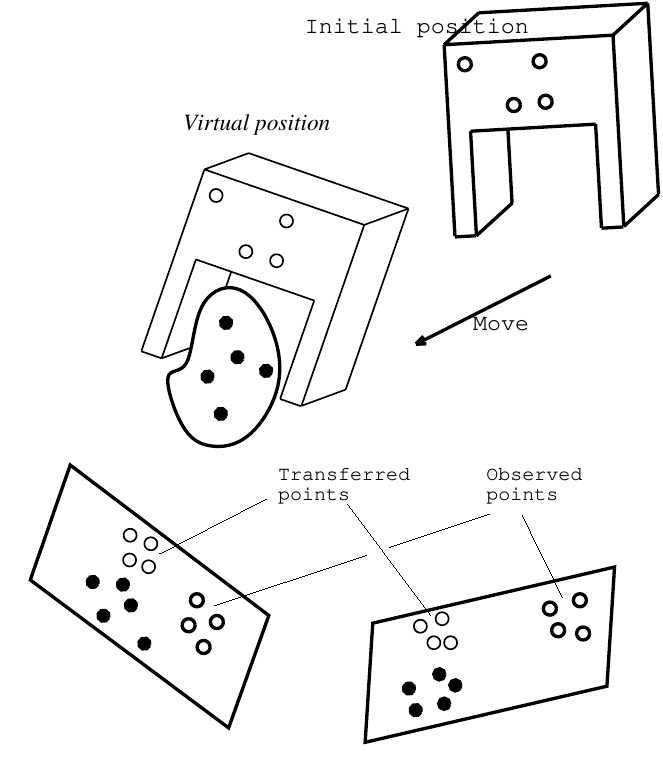}
\caption{This figure shows the runtime setup where the gripper is visually
servoed from an initial to a goal position. The goal position is defined by
the image transferred point, or implicitly by a 3-D virtual position of the
gripper.}
\label{fig:transfer}
\end{figure}

\subsection{Computing the set-point $\svect^{\star}$}

The set-point $\svect^{\star}$ is simply derived by transforming the 2-D
homogeneous coordinates of an image point into its image coordinates:
\[
\svect^{\star} = \left( \begin{array}{c} \widetilde{\bvect_1^x} \\
\vdots \\
\widetilde{\bvect_n^x} \end{array} \right)
\mbox{ with } \bvect_j^x = \lambda \left( \begin{array}{c} \widetilde{\bvect_1^x} \\ 1
\end{array} \right)
\] 

In theory the visual servoing algorithm described in
section~\ref{section:Vision-based-robot-motion} needs a single camera. Therefore
a minimal camera configuration may consist in one camera pair at planning and
a single camera at runtime: indeed, it is possible to combine the runtime
camera with any one of the two other cameras to perform the transfer and
compute the set-point. Alternatively, one can run two simultaneous visual
servoing processes and with two cameras eq.~(\ref{eq:control-velocity}) becomes:
\[
\Tvect_{g}
= g \left( \begin{array}{cc}
\hat{\Jmat}^\dag & \hat{\Jmat}'^\dag
\end{array} \right)
\left( \begin{array}{c}
\svect ^{\star} - \svect \\
\svect'^{\star} - \svect' 
 \end{array} \right)
\]

\section{Performance analysis}
\label{section:performance-analysis}

In this section we analyze the behavior of the visual servoing algorithm
described in section~\ref{section:Vision-based-robot-motion}. This algorithm is given an image 
set-point
$\svect^{\star}$ and a current image position $\svect$ and attempts to
align $\svect$ with $\svect^{\star}$. This alignment is done according to
eq.~(\ref{eq:control-velocity}): 
the robot moves until the norm of the image error vector
$\svect^{\star}-\svect$ vanishes. Therefore, a good estimation, 
$\hat{\Jmat}^\dag$, of the pseudo-inverse of $\Jmat$, is key. As already mentioned, the
classical approach used as an estimation of $\hat{\Jmat}$ is
the measured value of $\Jmat$ at the equilibrium configuration --- the robot lies
in the desired goal position. Hence, with this choice, $\hat{\Jmat}^\dag$ is kept constant during 
all the
servoing process.

The pose algorithm introduced in \cite{HoraudDornaikaLamiroyChristy97}
allows us to compute on-line a current estimate of
$\Jmat^\dag$ in approximatively $2\;10^{-3}$ seconds. This computation
time is compatible with real-time feature tracking and servoing. It is
therefore possible to run experiments in order to analyze the behavior of
visual servoing with an updated Jacobian.

Unlike the computation of the set-point $\svect^{\star}$,
both methods (updated and constant Jacobians) require explicit values for
the camera intrinsic parameters. However, in \cite{Espiau93} is shown that the
convergence of visual servoing is very little affected by these parameters. In
practice we used the horizontal and vertical focal lengths provided by the
camera manufacturer and we set the position of the optical axis at the image
center:
$\alpha_u = 1500, \;\; \alpha_v = 1000, \;\; u_0 = v_0 = 256$ 

\begin{figure}[t!]
\centering
\includegraphics[width=0.24\textwidth]{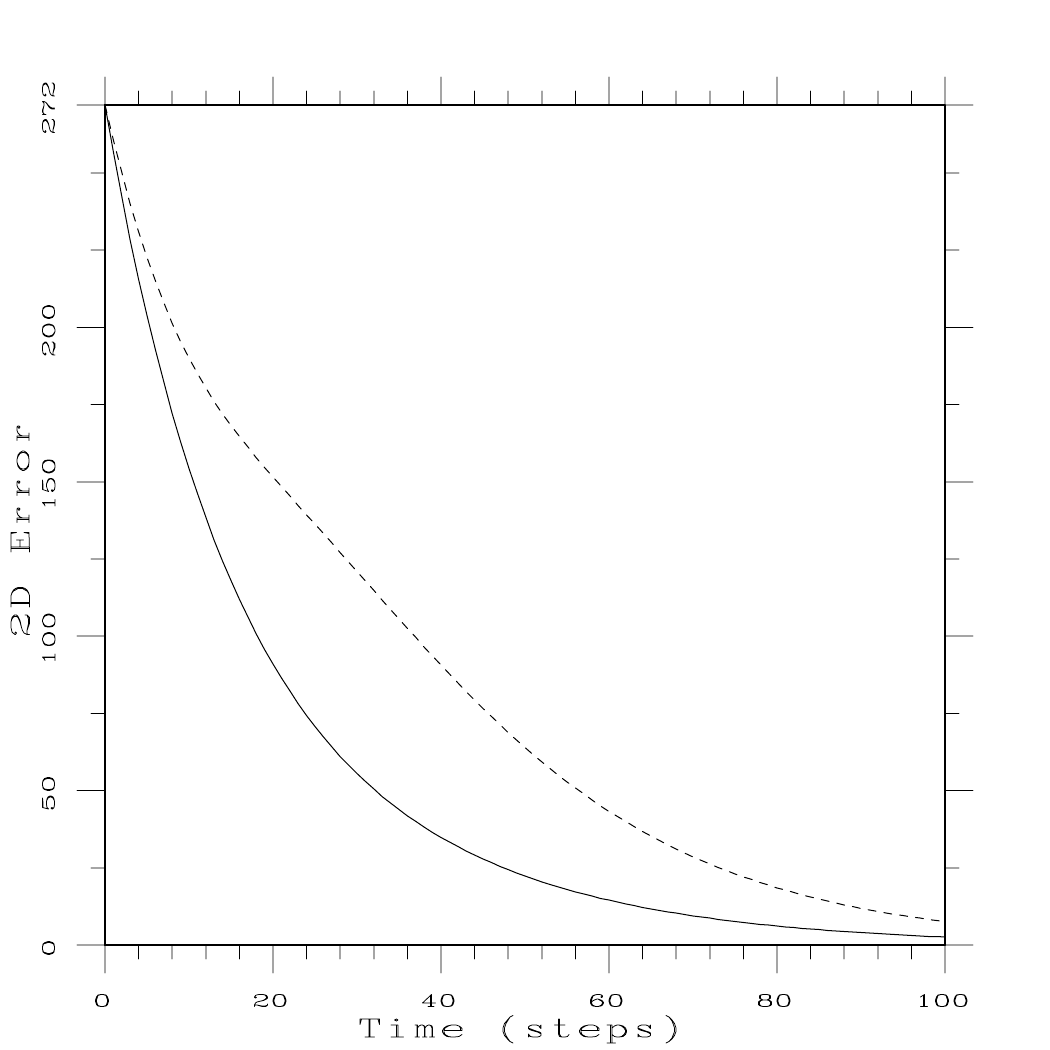}
\includegraphics[width=0.24\textwidth]{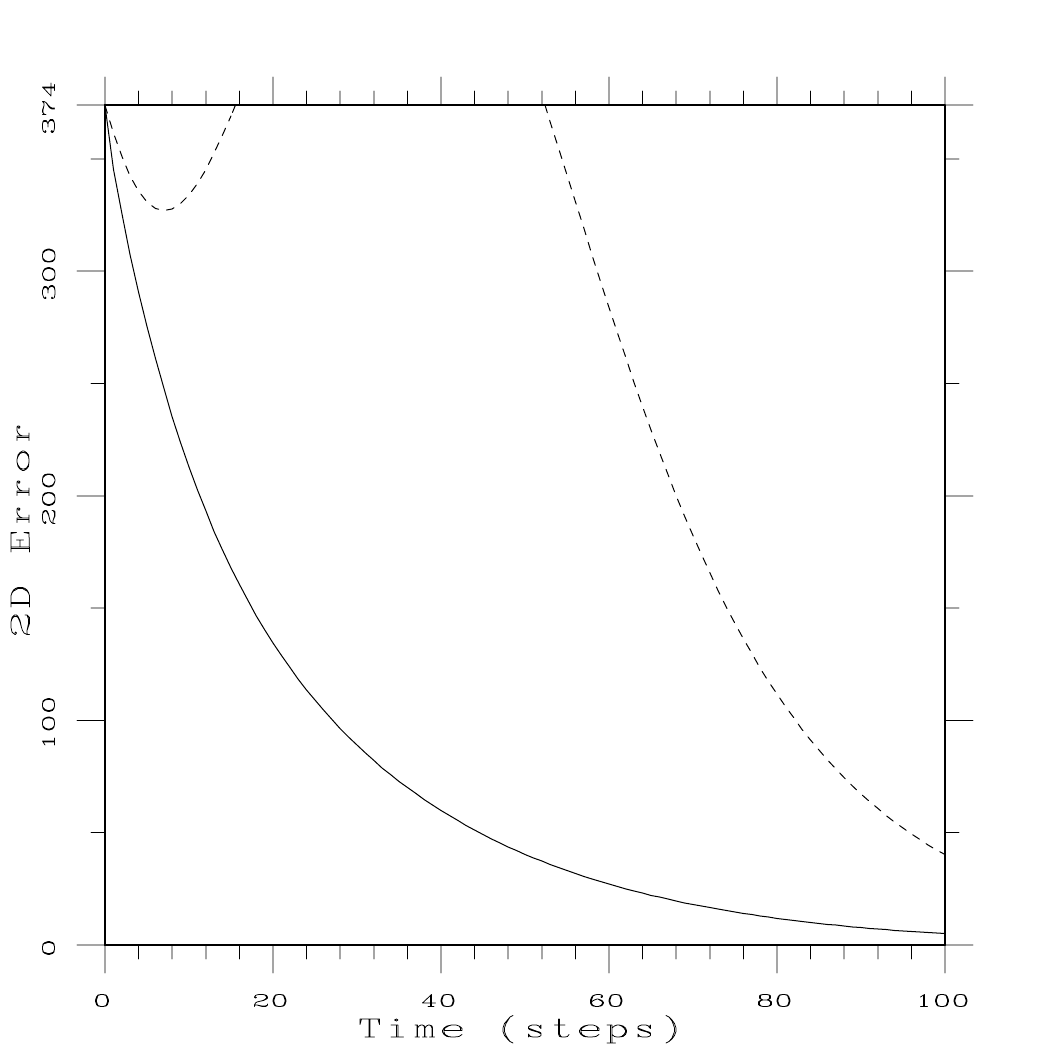}
\caption{These plots show the behavior of the servoing algorithm when the
distance between the initial and final gripper position is ``small" (left)
and when this distance is ``large" (right). The full
curves correspond to exact Jacobian servoing while the dashed curves
correspond to a constant Jacobian servoing.}
\label{fig:simulation-petit}
\end{figure}

In order to compare the behavior of the variable Jacobian servoing with the
constant Jacobian servoing we performed the following experiments. In the
first experiment the distance
between the initial and final robot position is ``small" (15$^0$ in
orientation and 35cm in depth). In the second experiment
this distance is large (30$^0$ in
orientation and 70cm in depth). The curves plotted on
Figure~\ref{fig:simulation-petit} represent the norm of the
image error ($\|\svect^{\star}-\svect\|$)
between the current gripper position and the final gripper
position as a function of time. 

One may notice that, in both experiments described above, 
the variable-Jacobian servoing
algorithm has an exponential error decrease associated with it, which is not
the case for the constant-Jacobian servoing and for large depth discrepancies
between the initial and goal positions.

The visual servoing algorithm runs at 10Hz on a Sun/Sparc10 workstation.
Table~\ref{table:servoing-times} summarizes the CPU times associated with each
stage of the algorithm. Notice that 70\% of the computing power is devoted to
data transfer (image acquisition, image transfer, computer-robot
communications) and only 2\% is devoted to the on-line computation of the
image
Jacobian.

\begin{table}[bhtp]
\label{table:servoing-times}
\caption{One cycle of the real-time control loop. }
\begin{center}
\begin{tabular}{||l|r||}
 \hline
Image acquisition                & 40ms  \\ \hline
Image transfer                  & 20ms  \\ \hline
Image processing                & 30ms  \\ \hline
Jacobian computation            & 2ms \\ \hline
Velocity screw computation      & 1ms \\ \hline
Computer-robot communication    & 10ms \\ \hline
{\bf Total }                    & 103ms \\ \hline
\end{tabular}
\end{center}
\end{table}

\section{Grasping experiments}
\label{section:grasping-experiments}

\begin{figure}[t!]
\centering
\frame{\includegraphics[width=0.24\textwidth]{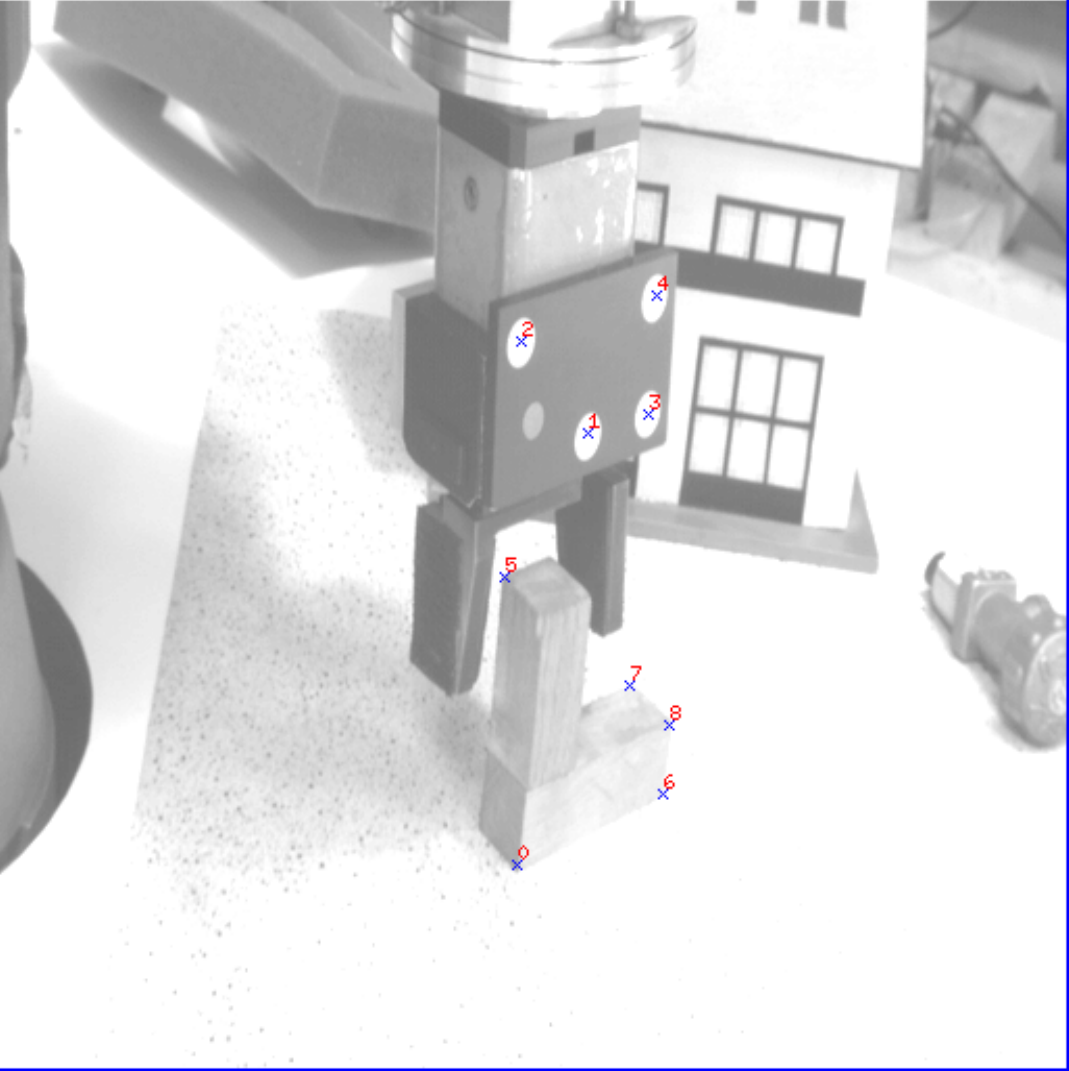}}
\frame{\includegraphics[width=0.24\textwidth]{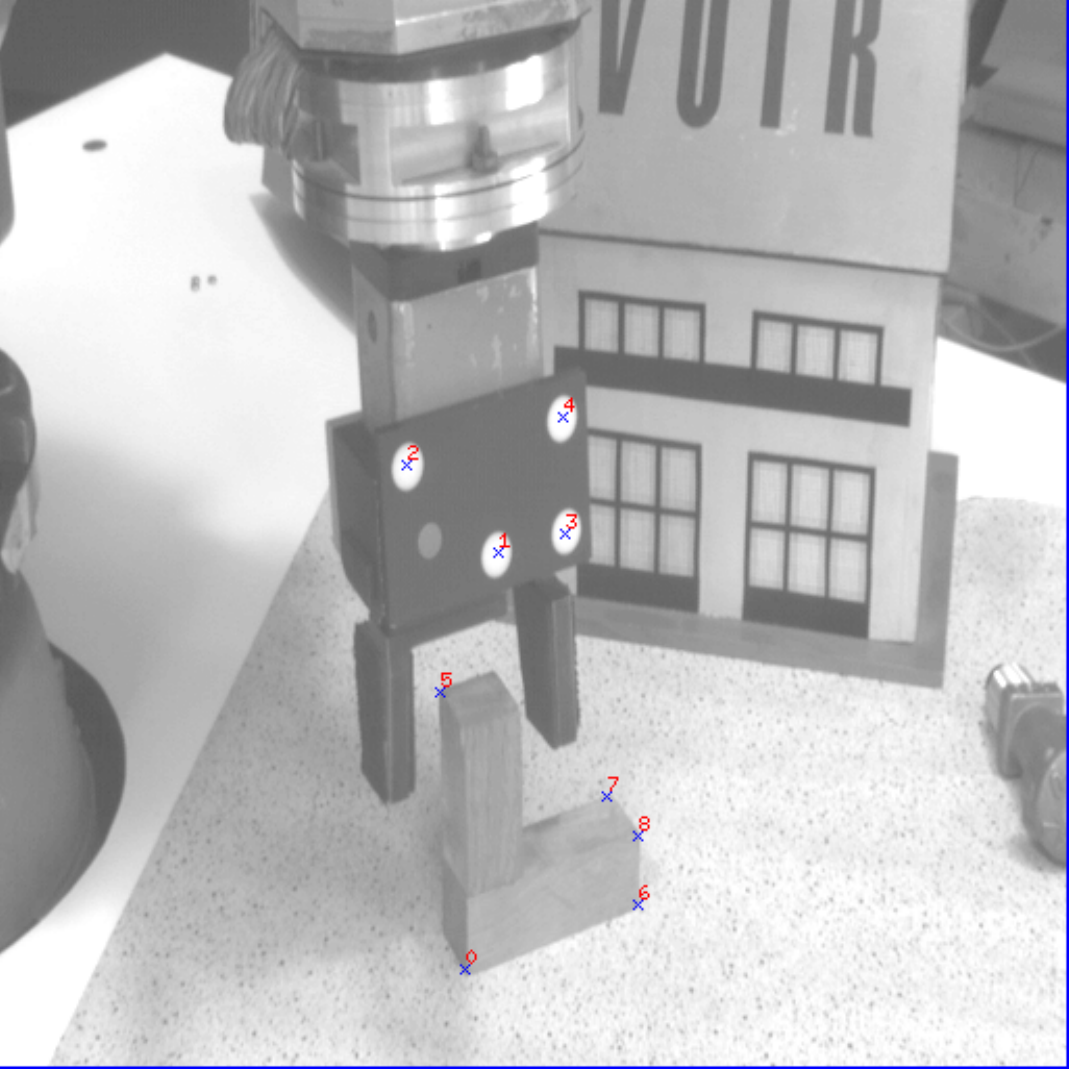}}
\caption{The gripper and the object to be grasped as viewed by a stereo rig. A
large set of point correspondences (not shown) allows us to compute the epipolar
geometry. Object points
together with gripper points are
represented in a 3-D projective space.}
\label{fig:setpoints}
\end{figure}

\begin{figure}[t!]
\centering
\frame{\includegraphics[width=0.24\textwidth]{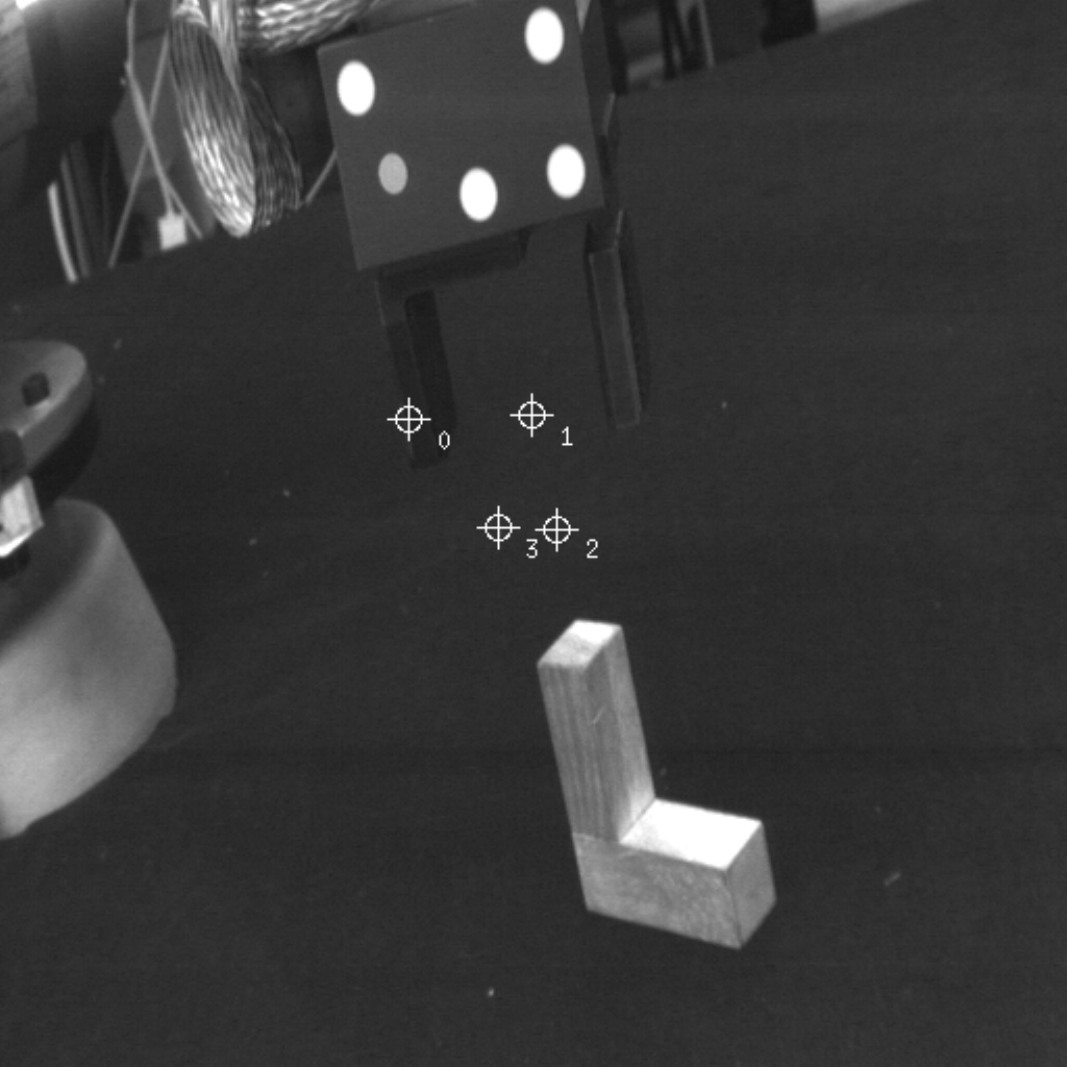}}
\frame{\includegraphics[width=0.24\textwidth]{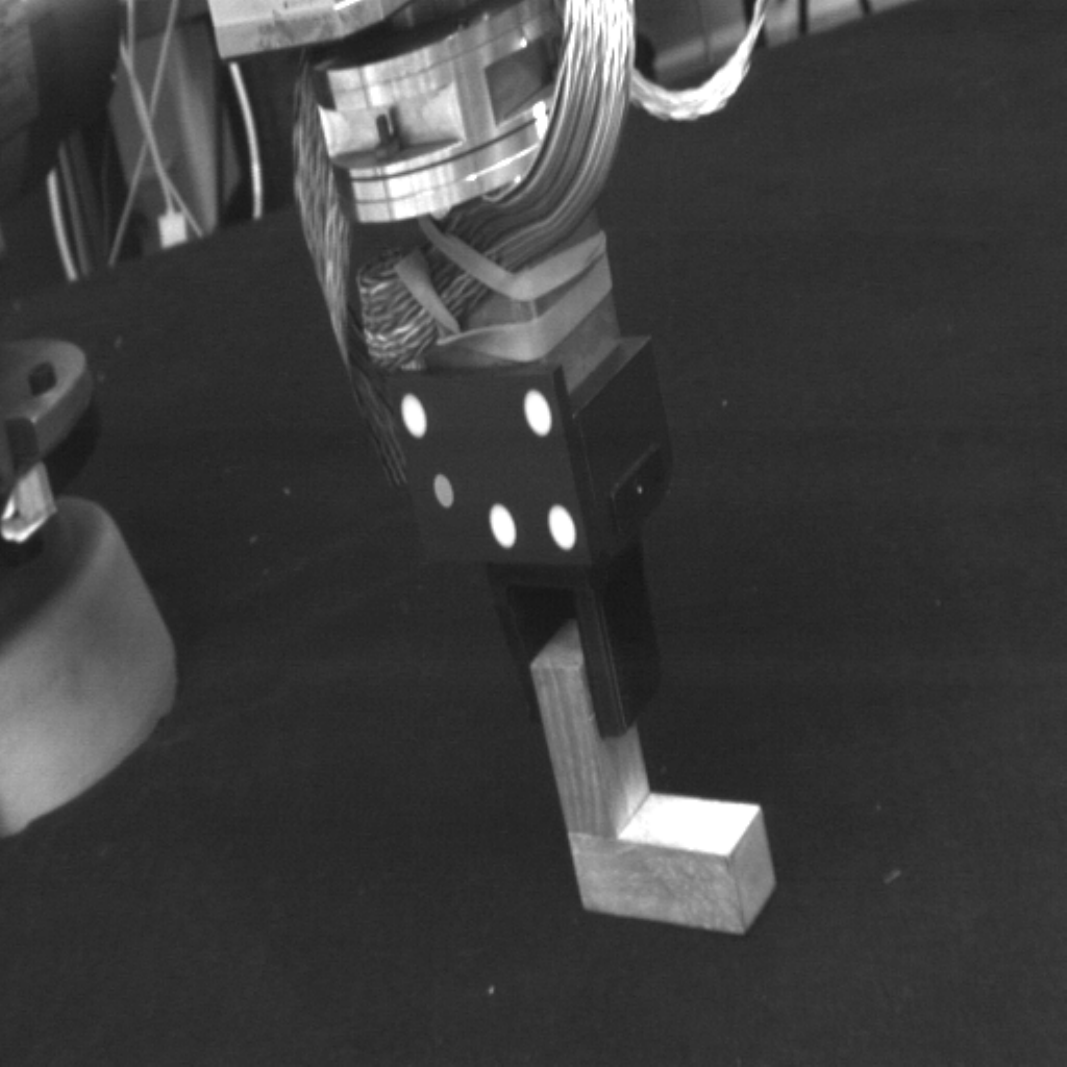}}
\caption{An example of applying the visually guided grasping method. The
set-point (left) is the projection of a view-invariant alignment
representation. The grasp (right) is reached when the image of the gripper is
aligned with the set-point.}
\label{fig:illustration}
\end{figure}

As already described, grasping includes a planning stage, a transfer stage,
and an execution
stage. The execution stage 
performs a real-time visually controlled 
loop:

\begin{itemize}
\item At planning time an uncalibrated stereo rig computes a 3-D
projective representation of grasping. This is illustrated on
Figure~\ref{fig:setpoints}. 

\item At preparation time a single camera observes both the object to be
grasped and the gripper in some initial position. The locations of both the
object and the gripper are arbitrary, provided that they are in the field of
view of the camera. The goal of this preparation stage is to
transfer gripper points in order to compute the
image set-point $\svect^\star$ -- Figure~\ref{fig:illustration}--left. 

\item The robot motion can now be controlled using visual feedback. The
velocity screw associated with the gripper frame is iteratively updated using
eq.~(\ref{eq:control-velocity}) until the norm of the image error vector
$\|\svect^\star - \svect\|$ vanishes. Figure~\ref{fig:illustration}--right 
shows the final grasping location reached by the gripper.
\end{itemize}
Since only one camera is used at runtime, the image point transfer technique
combines this camera with the camera pair used off-line to form two stereo
pairs.

One important feature of any grasping method is the precision with which the
gripper and the object are eventually aligned. In all our experiments the distance from
the camera to the object to be grasped is of approximatively 1 meter. The
camera lens has a focal length of 12.5 mm ($\alpha_u\approx 1000$) 
which allows for a
wide field of view. 
Since the method's main idea is to align image points, the final grasping
overall 
precision depends on the quality of the set-point $\svect^\star$. When the
gripper is properly aligned with the object to be grasped, a gripper point
with camera coordinates $(x, y, z)$ matches an image point with coordinates
$(u,v)$ and this image point belongs to the set-point $\svect^\star$. We
establish the relationship between the 3-D error and the 2-D error. 

By differentiation of eq.~(\ref{eq:u-hand-projection}) 
we obtain the following relationship:
\[
du  =  \alpha _u \left( \frac{dx}{z} - \frac{x \; dz}{z^2} \right) 
\]

The 3-D precision that we want to achieve is 0.5 mm. Therefore we have
$dx = dz = 5\; 10^{-4}$m, and let $x=0.1$m, $z=1$m, $\alpha_u=1000$. We
obtain: 
$du \approx dv \approx 0.5 \mbox{\rm pixels}$
This means that the transfer method outlined above must compute the set-point
with an accuracy of $0.5$ pixels. Such an accuracy may be obtained, provided
that (i)~the image locations of object points have an equivalent accuracy and
(ii)~there are 15 to 20 object points available with the image pairs
\cite{HoraudCsurka98}. The first condition can be easily satisfied with
standard correlation-based point-feature extraction methods. The second
condition is more difficult to satisfy because it is context dependent.

\section{Discussion}
\label{section:Discussion}
We described a method for aligning a robot end-effector with an object. An
example of such an alignment is grasping. The method consists of using an
uncalibrated stereo rig in order to represent the alignment in 3-D projective
space and of servoing the robot using visual feedback from either
one or two cameras. As
already mentioned, the set-point -- the set of image points with which the
gripper points must eventually be aligned -- can be computed without any
camera calibration. The final accuracy of the gripper-to-object alignment
depends on the accuracy with which the set-point has been estimated. 
Nevertheless, the computation of the image Jacobian
requires the camera intrinsic parameters to be known. The accuracy of 
these parameters does not
affect neither the final precision of the alignment nor the convergence of the
servoing algorithm; they merely affect the trajectory of the gripper between
its initial and goal locations.

One interesting feature of the method is that no Euclidean knowledge about the
object to be grasped is required. In order to relate the velocity screw
of the gripper with the image error vector the method requires Euclidean
knowledge about the robot gripper, namely the Euclidean coordinates of the
gripper markings must be known in gripper frame. This is an intrinsic property
of the gripper that can be easily determined using standard hand-eye or 
hand-tool calibration
methods \cite{HoraudDornaika95}, \cite{DornaikaHoraud98}. 

The use of visual feedback for object grasping and for alignment in general
is a promising research topic because it is a tolerant to various disturbances
and because it does not require such prior knowledge as robot-to-world
calibration and/or CAD models for the objects to be manipulated. The method
described in this paper permits a deeper understanding of the interaction
between uncalibrated vision and robot control which has important implications
in robotics. 

\balance

\end{document}